\PassOptionsToPackage{final}{graphicx}
\PassOptionsToPackage{final}{graphics}
\documentclass[letterpaper, 10 pt, conference]{ieeeconf}  

\IEEEoverridecommandlockouts                              

\usepackage{graphicx}
\usepackage{dblfloatfix}
\setkeys{Gin}{draft=false}
\usepackage{multirow}
\usepackage{amsmath} 
\usepackage{amssymb}  

\usepackage{marvosym}  

\title{
\large \bf
MRUF: Multi-granularity Routing with Uncertainty-Aware Fusion for Robust Multimodal Sentiment Analysis$\dagger$
}

\author{
Haoran Ma$^{1,2,3,4,5}$, Yinfeng Yu$^{1,2,3,4,5}$$^{,\mbox{\Letter}}$, and Liejun Wang$^{1,2,3,4,5}$%
\thanks{\small $\dagger$This work was supported in part by the National Natural Science Foundation of China under Grant Nos. 62463029 and 62472368.}%
\thanks{\small $^{1}$School of Computer Science and Technology, Xinjiang University, Urumqi 830017, China.}%
\thanks{\small $^{2}$Joint International Research Laboratory of Silk Road Multilingual Cognitive Computing.}%
\thanks{\small $^{3}$Xinjiang Multimodal Intelligent Processing and Information Security Engineering Technology Research Center.}%
\thanks{\small $^{4}$Pengcheng Laboratory Xinjiang Network Node.}%
\thanks{\small $^{5}$Embodied Intelligence Joint Laboratory.}%
\thanks{\small $^{\mbox{\Letter}}$Yinfeng Yu is the corresponding author (Email: yuyinfeng@xju.edu.cn).}%
}

\begin{document}

\maketitle
\thispagestyle{empty}
\pagestyle{empty}

\begin{abstract}
Multimodal sentiment analysis relies on language, visual, and acoustic cues, but utterance-level modality quality may vary due to occlusion, background noise, motion blur, or imperfect transcripts, causing conventional fusion to over-trust unreliable modalities. We propose MRUF, a reliability-aware fusion method that combines multi-granularity routing with uncertainty-aware calibration. MRUF summarizes sentiment-relevant representations, performs subspace- and modality-level routing, and supervises modality routing with leave-one-out error increases to estimate utterance-level modality importance. It further predicts modality-wise uncertainty and refines modality gates through inverse-variance reweighting, while modality-invariant contrastive alignment stabilizes the shared representation space. Experiments on CMU-MOSI and CMU-MOSEI under aligned and unaligned settings show consistent improvements over strong baselines, and mechanism analysis verifies that modalities with higher predicted uncertainty receive lower fusion weights.
\end{abstract}

\section{Introduction}

Multimodal sentiment analysis predicts sentiment by jointly modeling language, visual, and acoustic cues~\cite{baltruaitis2019multimodal,zadeh2016multimodal,zadeh2018multimodal}. It is important for human-centered intelligent systems, such as human--machine interaction, driver monitoring, intelligent tutoring, and affective decision support. In practice, multimodal signals are often asynchronous, heterogeneous, and uneven in quality: visual streams may suffer from occlusion, motion blur, or head-pose changes; acoustic signals may be affected by background noise or channel variation; and transcripts may be incomplete or inaccurate. Such quality fluctuations can make a model over-trust unreliable evidence and produce unstable predictions.

Existing studies mainly improve multimodal sentiment analysis through cross-modal interaction and robust representation learning. Fusion methods capture modality correlations via tensor fusion, low-rank fusion, attention, graph interaction, progressive alignment, or interaction enhancement~\cite{zadeh2017tensor,liu2018efficient,tsai2019multimodal,yang2021mtag,lv2021progressive,liang2021attention}, while robust methods handle missing or degraded modalities through calibration, contrastive learning, reconstruction, dynamic weighting, expert modeling, or modality-invariant distillation~\cite{tu2024multimodal,liu2024contrastive,peng2024carat,fang2025emoe,wang2025modality}. Similar reliability concerns also appear in audio-visual navigation, vision-and-language navigation, audio-visual source separation, speech processing, and visual detail enhancement~\cite{YinfengICLR2022saavn,yu2023measuring,li2025audio,yang2026beyond,yu2025dope,yu2025dgfnet}. However, modality importance is often learned implicitly, and uncertainty is rarely used to directly calibrate modality weights during final fusion.

These issues remain even with strong representation backbones. Disentangled or decoupled models improve representation quality by separating modality-invariant and modality-specific factors or transferring knowledge across modalities~\cite{hazarika2020misa,li2023decoupled}. Nevertheless, a degraded modality may still receive excessive influence if final fusion is not guided by task contribution and uncertainty.

To this end, we propose MRUF (Multi-granularity Routing with Uncertainty-Aware Fusion), a reliability-aware multimodal fusion method built on the decoupling-distillation backbone DMD~\cite{li2023decoupled}. MRUF performs subspace- and modality-level routing, supervises modality routing with leave-one-out error increases, and calibrates modality gates through uncertainty-based inverse-variance reweighting. Modality-invariant contrastive alignment is further used to stabilize the shared representation space.

Experiments on CMU-MOSI and CMU-MOSEI under aligned and unaligned settings show that MRUF consistently improves over strong baselines. Further analysis shows that modalities with higher predicted uncertainty tend to receive lower fusion weights, supporting the intended reliability-aware behavior.

The main contributions of this paper are as follows:
\begin{itemize}
\item We propose MRUF, a reliability-aware multimodal fusion method built upon a decoupling-distillation backbone.

\item We design task-aware multi-granularity routing to estimate interpretable utterance-level modality importance.

\item We introduce uncertainty-aware gate calibration and modality-invariant contrastive alignment to improve robustness under modality quality variations.
\end{itemize}

\section{Related Work}

\subsection{Multimodal Sentiment Analysis}

Multimodal sentiment analysis has evolved from explicit fusion to cross-modal interaction and robust multimodal learning. Early methods, such as TFN~\cite{zadeh2017tensor} and LMF~\cite{liu2018efficient}, model high-order or low-rank cross-modal correlations, while later methods improve asynchronous multimodal modeling through attention, graph interaction, progressive relation modeling, or interaction enhancement~\cite{tsai2019multimodal,yang2021mtag,lv2021progressive,liang2021attention}. Recent studies further explore refined co-space representation learning, audiovisual sentiment modeling, nonverbal behavior-enhanced language representations, asynchronous modality reinforcement, and modality-invariant temporal representation distillation~\cite{shi2024co,yu2021weavenet,wang2019words,yang2023target,wang2025modality}. These methods improve multimodal interaction and alignment, but modality reliability is usually inferred implicitly.

Related multimodal perception and speech-oriented studies also emphasize reliable cross-modal modeling, including audio-visual navigation and acoustic perception~\cite{YinfengICLR2022saavn,yu2023measuring,li2025audio,zhang2025iterative,zhang2025advancing,yu2025dynamic}, vision-and-language navigation~\cite{yang2026beyond,yu2025dope}, audio-visual source separation and speech processing~\cite{yu2025dgfnet,mattursun2024bss,zhang2024nonlinear,cao2024vnet}, and visual detail enhancement~\cite{fu2025fsdenet}. These studies similarly motivate adaptive use of heterogeneous modalities.

Robust multimodal learning handles missing, noisy, or corrupted modalities through modality dropping, gated fusion, memory-based fusion, calibration-aware learning, feature restoration, contrastive learning, reconstruction, or expert modeling~\cite{neverova2015moddrop,arevalo2017gated,zadeh2018memory,tu2024multimodal,sun2023efficient,liu2024contrastive,peng2024carat,fang2025emoe}. Unlike these methods, MRUF explicitly supervises utterance-level modality routing with task-aware leave-one-out error increases and calibrates final fusion gates using predicted modality uncertainty.

\subsection{Disentangled Learning and Decoupled Distillation}

Disentangled multimodal learning separates modality-invariant and modality-specific factors to obtain more robust representations~\cite{tsai2018learning,yang2022disentangled}. MISA~\cite{hazarika2020misa} learns shared and private representations, while cyclic translation-based methods enhance shared representations under modality shift~\cite{pham2019found}. DMD~\cite{li2023decoupled} further combines representation decoupling with graph distillation. However, these methods do not explicitly enforce utterance-level modality weights to reflect task contribution or uncertainty. MRUF builds on this line of work by adding temporal summarization, task-aware multi-granularity routing, modality-invariant contrastive alignment, and uncertainty-aware gate calibration.

\section{The Proposed Method}

\begin{figure*}[!t]
\centering
\includegraphics[width=0.8\textwidth]{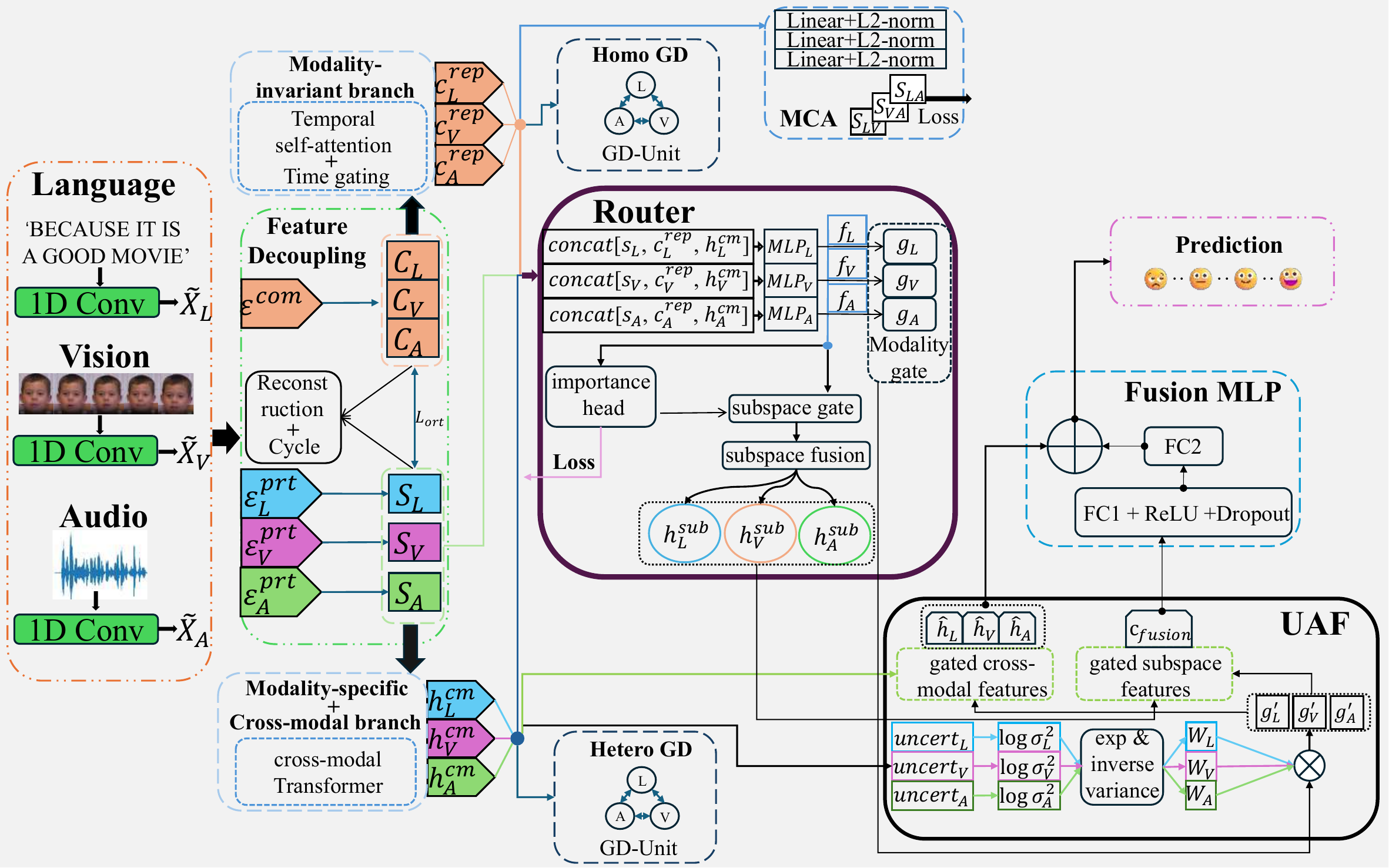}
\caption{Overview of MRUF. MRUF extends DMD representations with invariant-branch summarization, modality-invariant contrastive alignment, task-aware multi-granularity routing, and uncertainty-aware fusion. Leave-one-out supervision is used only during training.}
\label{fig:framework}
\end{figure*}

Multimodal sentiment analysis requires integrating heterogeneous cues with varying reliability. Although DMD provides decoupled modality-invariant and modality-specific representations with cross-modal distillation, its final fusion stage may still over-trust degraded modalities. MRUF addresses this issue by estimating utterance-level modality importance and calibrating it with predicted uncertainty.

As shown in Fig.~\ref{fig:framework}, MRUF contains invariant-branch temporal summarization, modality-invariant contrastive alignment (MCA), multi-granularity routing, and uncertainty-aware fusion (UAF). Let $\mathcal{M}=\{L,V,A\}$ denote language, vision, and audio. Following DMD, each modality is decomposed into a modality-specific sequence $\mathbf{S}_m=[\mathbf{s}_{m,1},\ldots,\mathbf{s}_{m,T_m}]$ and a modality-invariant sequence $\mathbf{C}_m=[\mathbf{c}_{m,1},\ldots,\mathbf{c}_{m,T_m}]$, with a cross-modal summary $\mathbf{h}^{\mathrm{cm}}_m$. MRUF performs reliability-aware fusion on these utterance-level representations.

\subsection{Invariant-Branch Temporal Summarization}

The invariant branch contains shared semantic information, but different temporal positions contribute unequally to sentiment prediction. MRUF applies temporal gating and self-attentive aggregation to obtain an utterance-level invariant representation:
\begin{equation}
\begin{aligned}
q_{m,t} &= \sigma(\mathbf{w}_{m}^{\top}\mathbf{c}_{m,t}+b_m),\\
\mathbf{c}^{\mathrm{time}}_m
&=
\frac{\sum_{t=1}^{T_m} q_{m,t}\mathbf{c}_{m,t}}
{\sum_{t=1}^{T_m} q_{m,t}+\epsilon},\\
\mathbf{c}^{\mathrm{rep}}_m
&=
\frac{1}{2}
\left(
\mathbf{c}^{\mathrm{time}}_m+\mathbf{c}^{\mathrm{att}}_m
\right),
\end{aligned}
\end{equation}
where $\mathbf{c}^{\mathrm{att}}_m$ is obtained by applying a self-attention encoder to the invariant sequence and averaging the last-layer outputs.

\subsection{Modality-Invariant Contrastive Alignment}

To stabilize the shared semantic space used by routing, MRUF introduces MCA as an auxiliary regularizer. Each invariant summary $\mathbf{c}^{\mathrm{rep}}_{m,i}$ is projected and $\ell_2$-normalized as $\mathbf{z}_{m,i}$. For an ordered modality pair $(m,n)$, the contrastive loss is
\begin{equation}
\ell_{m\rightarrow n}
=
-\frac{1}{N}\sum_{i=1}^{N}
\log
\frac{
\exp(\mathbf{z}_{m,i}^{\top}\mathbf{z}_{n,i}/\tau)
}{
\sum_{j=1}^{N}\exp(\mathbf{z}_{m,i}^{\top}\mathbf{z}_{n,j}/\tau)
},
\end{equation}
where $\tau$ is a learnable temperature. The final MCA loss averages all six cross-modal directions:
\begin{equation}
\mathcal{L}_{\mathrm{MCA}}
=
\frac{1}{6}\sum_{m\neq n}\ell_{m\rightarrow n},
\quad m,n\in\mathcal{M}.
\end{equation}

\subsection{Multi-granularity Routing}

MRUF performs routing at both the subspace level and the modality level. For each modality, the specific sequence is mean-pooled to obtain $\mathbf{s}^{\mathrm{rep}}_m$. The router input combines modality-specific, modality-invariant, and cross-modal information:
\begin{equation}
\mathbf{f}_m=
\mathrm{MLP}^{r}_m
\left(
[\mathbf{s}^{\mathrm{rep}}_m;\mathbf{c}^{\mathrm{rep}}_m;\mathbf{h}^{\mathrm{cm}}_m]
\right).
\end{equation}

\subsubsection{Subspace Routing}

Subspace routing balances modality-specific and modality-invariant information within each modality:
\begin{equation}
\begin{aligned}
\mathbf{g}^{\mathrm{sub}}_m
&=
\mathrm{softmax}
\left(
\mathbf{W}^{\mathrm{sub}}_m\mathbf{f}_m+\mathbf{b}^{\mathrm{sub}}_m
\right)
=
[g^{\mathrm{sub}}_{m,s},g^{\mathrm{sub}}_{m,c}],\\
\mathbf{h}^{\mathrm{sub}}_m
&=
g^{\mathrm{sub}}_{m,s}\mathbf{s}^{\mathrm{rep}}_m+
g^{\mathrm{sub}}_{m,c}\mathbf{c}^{\mathrm{rep}}_m .
\end{aligned}
\end{equation}

\subsubsection{Modality Routing}

Modality routing estimates utterance-level modality importance. MRUF predicts an independent raw modality gate and an auxiliary normalized importance distribution:
\begin{equation}
\begin{aligned}
g^{\mathrm{mod}}_m
&=
\sigma(\mathbf{w}^{\mathrm{mod}\top}_m\mathbf{f}_m+b^{\mathrm{mod}}_m),\\
\hat{\boldsymbol{\pi}}
&=
\mathrm{softmax}
\left(
\mathrm{MLP}^{\mathrm{imp}}
([\mathbf{f}_L;\mathbf{f}_V;\mathbf{f}_A])
\right).
\end{aligned}
\end{equation}

\subsubsection{Task-aware Leave-one-out Supervision}

To make modality routing reflect task contribution, MRUF constructs a leave-one-out teacher during training. Let $\hat{y}^{\mathrm{all}}$ be the prediction with all modalities, and let $\hat{y}^{\setminus m}$ be the prediction obtained by masking modality $m$. The contribution of modality $m$ is measured by the error increase:
\begin{equation}
\Delta_m=
\max
\left(
\left|\hat{y}^{\setminus m}-y\right|
-
\left|\hat{y}^{\mathrm{all}}-y\right|,
0
\right).
\end{equation}
The normalized teacher importance and router supervision loss are
\begin{equation}
\tilde{I}_m=
\frac{\Delta_m}{\sum_{k\in\mathcal{M}}\Delta_k+10^{-8}},
\quad
\mathcal{L}_{\mathrm{router}}=
\frac{1}{3}
\sum_{m\in\mathcal{M}}
(\hat{\pi}_m-\tilde{I}_m)^2 .
\end{equation}
This supervision is used only during training. At inference time, MRUF does not require masked forward passes and uses only one forward pass.

\subsection{Uncertainty-Aware Fusion}

Content-based routing may still be unreliable when a modality is severely corrupted. Inspired by uncertainty-based weighting strategies~\cite{kendall2017uncertainties}, MRUF predicts a modality-wise log-variance proxy and converts it into a normalized inverse-variance weight:
\begin{equation}
\begin{aligned}
\log \sigma_m^2
&=
\mathbf{w}^{u\top}_m\mathbf{h}^{\mathrm{cm}}_m+b^u_m,\\
\omega_m
&=
\frac{
(\sigma_m^2+\epsilon)^{-1}
}{
\sum_{k\in\mathcal{M}}(\sigma_k^2+\epsilon)^{-1}
}.
\end{aligned}
\end{equation}
The final reliability-aware modality gate is
\begin{equation}
g_m=g^{\mathrm{mod}}_m\omega_m.
\end{equation}
Thus, a modality with higher predicted uncertainty receives a smaller calibration weight and contributes less to final fusion.

\subsection{Reliability-Aware Prediction and Training Objective}

MRUF uses the uncertainty-calibrated gate $g_m$ to modulate both cross-modal and subspace-routed features:
\begin{equation}
\bar{\mathbf{h}}^{\mathrm{cm}}_m=
g_m\,\sigma(\mathbf{W}^{\mathrm{cm}}_m\mathbf{h}^{\mathrm{cm}}_m),
\quad
\bar{\mathbf{h}}^{\mathrm{sub}}_m=
g_m\,\mathbf{h}^{\mathrm{sub}}_m .
\end{equation}
The gated modality features are then concatenated and fed into a residual MLP for prediction:
\begin{equation}
\hat{y}
=
\mathrm{MLP}_{\mathrm{res}}
\left(
[\bar{\mathbf{h}}^{\mathrm{cm}}_L;
 \bar{\mathbf{h}}^{\mathrm{cm}}_V;
 \bar{\mathbf{h}}^{\mathrm{cm}}_A;
 \mathbf{c}^{\mathrm{fus}}]
\right),
\end{equation}
where $\mathbf{c}^{\mathrm{fus}}$ is obtained by projecting the concatenated gated subspace features.

The model is optimized jointly with the DMD backbone. Let $\mathcal{L}_{\mathrm{task}}$ denote the prediction loss, $\mathcal{L}_{\mathrm{dist}}$ denote the graph-distillation loss, and $\mathcal{L}_{\mathrm{reg}}$ collect the regularizers inherited from DMD. The overall objective is
\begin{equation}
\mathcal{L}
=
\mathcal{L}_{\mathrm{task}}
+\mathcal{L}_{\mathrm{dist}}
+0.1\,\mathcal{L}_{\mathrm{reg}}
+\lambda_r\mathcal{L}_{\mathrm{router}}
+\lambda_m\mathcal{L}_{\mathrm{MCA}}.
\end{equation}
Compared with DMD, MRUF introduces additional training cost because the leave-one-out teacher requires masked forward passes. This cost is removed at inference, where the model still uses a single forward pass.

\section{Experiments}

\begin{table}[t]
\centering
\caption{Comparison on CMU-MOSI. ACC$_7$, ACC$_2$, and F1 are percentages. ``*'' denotes BERT-based language features~\cite{devlin2019bert}. Results are averaged over three seeds.}
\label{tab:mosi}
\footnotesize
\setlength{\tabcolsep}{4pt}
\renewcommand{\arraystretch}{1.05}
\begin{tabular}{l|c|cccc}
\hline
Method & Setting & ACC$_7$ $\uparrow$ & ACC$_2$ $\uparrow$ & F1 $\uparrow$ & MAE $\downarrow$ \\
\hline
TFN~\cite{zadeh2017tensor}          & \multirow{11}{*}{Aligned}   & 30.4 & 76.2 & 76.3 & 0.974 \\
LMF~\cite{liu2018efficient}         &                             & 34.5 & 77.2 & 77.2 & 0.951 \\
MulT~\cite{tsai2019multimodal}      &                             & 33.9 & 78.7 & 78.9 & 0.939 \\
RAVEN~\cite{wang2019words}          &                             & 33.8 & 76.5 & 76.6 & 0.913 \\
MICA~\cite{liang2021attention}      &                             & 35.2 & 78.7 & 78.8 & 0.909 \\
MTAG~\cite{yang2021mtag}            &                             & 31.1 & 81.8 & 81.8 & 0.929 \\
PMR~\cite{lv2021progressive}        &                             & 34.1 & 75.4 & 75.5 & 1.024 \\
Self-MM*~\cite{yu2021learning}      &                             & 43.9 & 83.5 & 83.5 & 0.742 \\
EMOE*~\cite{fang2025emoe}           &                             & 44.7 & 83.9 & 83.9 & 0.731 \\
DMD*~\cite{li2023decoupled}         &                             & 44.5 & 83.2 & 83.3 & 0.729 \\
\textbf{MRUF (Ours)*}               &                             & \textbf{46.7} & \textbf{84.5} & \textbf{84.4} & \textbf{0.717} \\
\hline
TFN~\cite{zadeh2017tensor}          & \multirow{11}{*}{Unaligned} & 33.2 & 75.1 & 75.2 & 0.982 \\
LMF~\cite{liu2018efficient}         &                             & 30.5 & 77.8 & 77.9 & 0.967 \\
MulT~\cite{tsai2019multimodal}      &                             & 32.4 & 79.0 & 79.0 & 0.929 \\
RAVEN~\cite{wang2019words}          &                             & 33.9 & 75.4 & 75.5 & 0.917 \\
MICA~\cite{liang2021attention}      &                             & 34.7 & 78.8 & 78.9 & 0.916 \\
MTAG~\cite{yang2021mtag}            &                             & 30.1 & 81.8 & 81.9 & 0.936 \\
PMR~\cite{lv2021progressive}        &                             & 34.3 & 76.4 & 76.5 & 0.995 \\
Self-MM*~\cite{yu2021learning}      &                             & 44.1 & 82.1 & 82.2 & 0.732 \\
EMOE*~\cite{fang2025emoe}           &                             & 44.8 & 83.8 & 83.9 & 0.722 \\
DMD*~\cite{li2023decoupled}         &                             & 44.7 & 83.3 & 83.3 & 0.721 \\
\textbf{MRUF (Ours)*}               &                             & \textbf{46.5} & \textbf{84.4} & \textbf{84.3} & \textbf{0.701} \\
\hline
\end{tabular}
\end{table}

\begin{table}[t]
\centering
\caption{Comparison on CMU-MOSEI. ACC$_7$, ACC$_2$, and F1 are percentages. ``*'' denotes BERT-based language features~\cite{devlin2019bert}. Results are averaged over three seeds.}
\label{tab:mosei}
\footnotesize
\setlength{\tabcolsep}{4pt}
\renewcommand{\arraystretch}{1.05}
\begin{tabular}{l|c|cccc}
\hline
Method & Setting & ACC$_7$ $\uparrow$ & ACC$_2$ $\uparrow$ & F1 $\uparrow$ & MAE $\downarrow$ \\
\hline
TFN~\cite{zadeh2017tensor}          & \multirow{11}{*}{Aligned}   & 49.5 & 79.1 & 79.2 & 0.584 \\
LMF~\cite{liu2018efficient}         &                             & 50.9 & 82.8 & 82.7 & 0.567 \\
MulT~\cite{tsai2019multimodal}      &                             & 50.8 & 81.1 & 81.2 & 0.625 \\
RAVEN~\cite{wang2019words}          &                             & 49.2 & 80.3 & 80.3 & 0.572 \\
MICA~\cite{liang2021attention}      &                             & 48.8 & 80.6 & 80.7 & 0.583 \\
MTAG~\cite{yang2021mtag}            &                             & 50.5 & 82.5 & 82.6 & 0.559 \\
PMR~\cite{lv2021progressive}        &                             & 48.4 & 78.6 & 78.7 & 0.663 \\
Self-MM*~\cite{yu2021learning}      &                             & 52.1 & 83.1 & 83.1 & 0.546 \\
EMOE*~\cite{fang2025emoe}           &                             & 52.2 & 83.9 & 83.9 & 0.549 \\
DMD*~\cite{li2023decoupled}         &                             & 51.6 & 83.5 & 83.5 & 0.553 \\
\textbf{MRUF (Ours)*}               &                             & \textbf{52.3} & \textbf{84.5} & \textbf{84.4} & \textbf{0.544} \\
\hline
TFN~\cite{zadeh2017tensor}          & \multirow{11}{*}{Unaligned} & 49.1 & 82.3 & 82.4 & 0.572 \\
LMF~\cite{liu2018efficient}         &                             & 50.5 & 82.2 & 82.3 & 0.559 \\
MulT~\cite{tsai2019multimodal}      &                             & 51.5 & 82.1 & 82.2 & 0.545 \\
RAVEN~\cite{wang2019words}          &                             & 50.2 & 81.5 & 81.4 & 0.562 \\
MICA~\cite{liang2021attention}      &                             & 49.4 & 81.1 & 81.0 & 0.583 \\
MTAG~\cite{yang2021mtag}            &                             & 51.2 & 83.2 & 81.1 & 0.567 \\
PMR~\cite{lv2021progressive}        &                             & 48.9 & 79.3 & 79.1 & 0.634 \\
Self-MM*~\cite{yu2021learning}      &                             & 51.9 & 83.3 & 83.1 & 0.539 \\
EMOE*~\cite{fang2025emoe}           &                             & 52.2 & 84.1 & 84.0 & 0.547 \\
DMD*~\cite{li2023decoupled}         &                             & 52.1 & 83.9 & 83.8 & 0.542 \\
\textbf{MRUF (Ours)*}               &                             & \textbf{52.7} & \textbf{84.5} & \textbf{84.4} & \textbf{0.538} \\
\hline
\end{tabular}
\end{table}

\subsection{Datasets}

We evaluate MRUF on CMU-MOSI~\cite{zadeh2016multimodal} and CMU-MOSEI~\cite{zadeh2018multimodal}. CMU-MOSI contains 2,199 utterances with the standard train/validation/test split of 1,284/229/686, while CMU-MOSEI contains 22,856 utterances with the conventional split of 16,326/1,871/4,659. Each utterance is annotated with a sentiment score in $[-3,3]$.

Following prior work, we report results under both aligned and unaligned settings. The aligned setting uses temporally synchronized multimodal features, while the unaligned setting keeps the original asynchronous sequences. We use BERT-based textual features, FACET-based visual features, and COVAREP acoustic features.

\subsection{Evaluation Metrics}

Following prior studies~\cite{li2023decoupled,fang2025emoe}, we report ACC$_7$, ACC$_2$, F1, and MAE. ACC$_7$ evaluates seven-class sentiment classification, while ACC$_2$ and F1 evaluate binary sentiment polarity by treating scores smaller than $0$ as negative and the others as non-negative. Higher ACC$_7$, ACC$_2$, and F1 indicate better classification performance, while lower MAE indicates better regression quality.

\subsection{Implementation Details}

All models are implemented in PyTorch and trained on a single NVIDIA Tesla T4 GPU with 16\,GB memory. We use a batch size of 16 and early stopping with a patience of 10 epochs. The checkpoint with the best validation performance is evaluated on the test set, and all results are averaged over three random seeds. We set $\lambda_r=0.3$ and $\lambda_m=0.07$ for the router supervision loss and the modality-invariant contrastive alignment loss, respectively.

\subsection{Efficiency Analysis}

MRUF introduces additional training cost because leave-one-out router supervision requires modality-masked forward passes to estimate task-aware modality contribution. This cost is used only during training and is removed at inference, where MRUF still uses a single forward pass. On CMU-MOSEI under the aligned setting, MRUF requires 539.3\,s per training epoch, about 150.7\,ms per test batch in the evaluation pipeline, and contains 112.40\,M parameters. These results indicate that the additional cost mainly appears in training, while the inference procedure remains a single-forward-pass process.

\begin{figure*}[!t]
\centering
\begin{minipage}[t]{0.32\textwidth}
    \centering
    \includegraphics[width=0.8\textwidth]{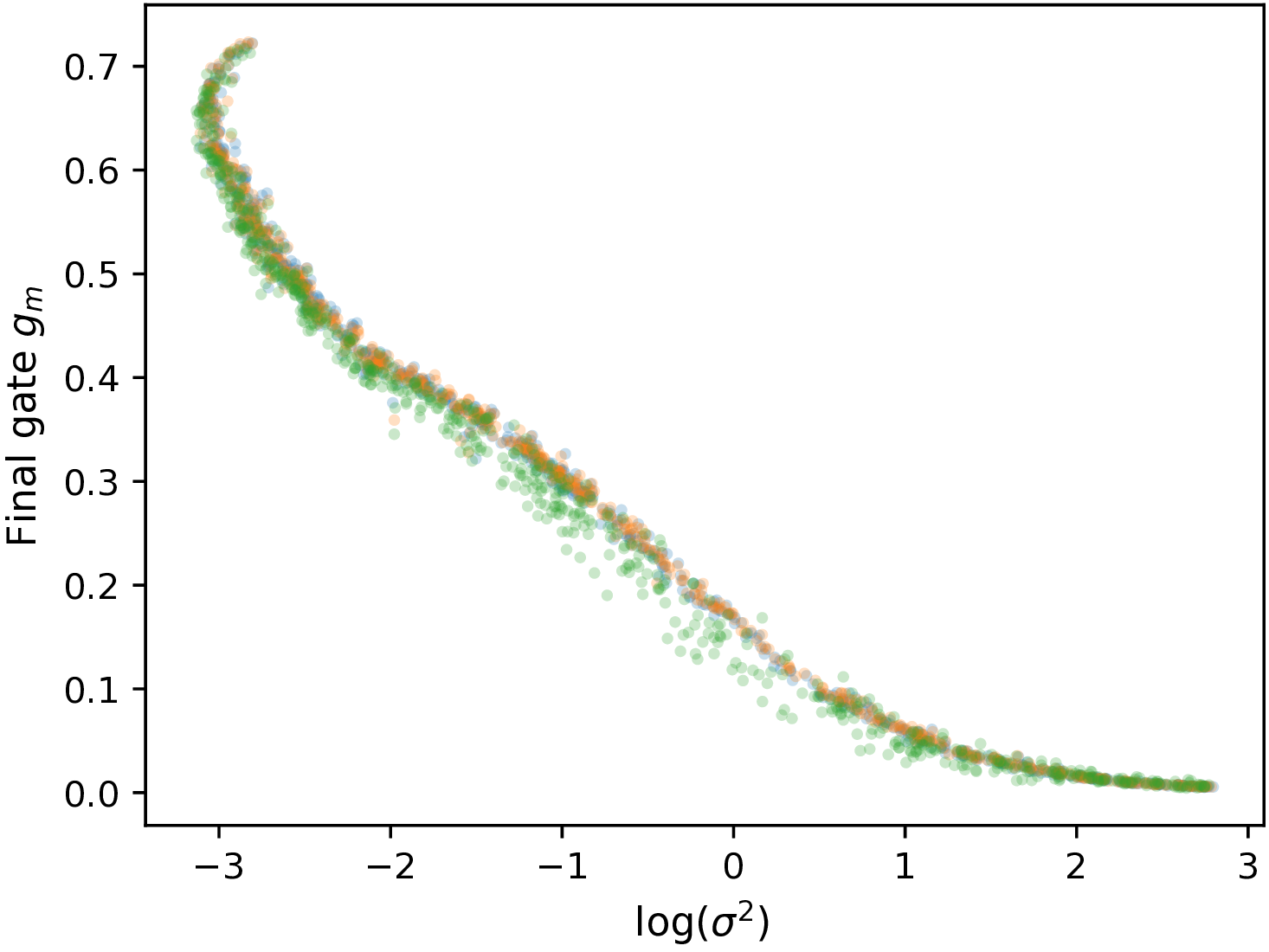}

    (a) Language
\end{minipage}
\hfill
\begin{minipage}[t]{0.32\textwidth}
    \centering
    \includegraphics[width=0.8\textwidth]{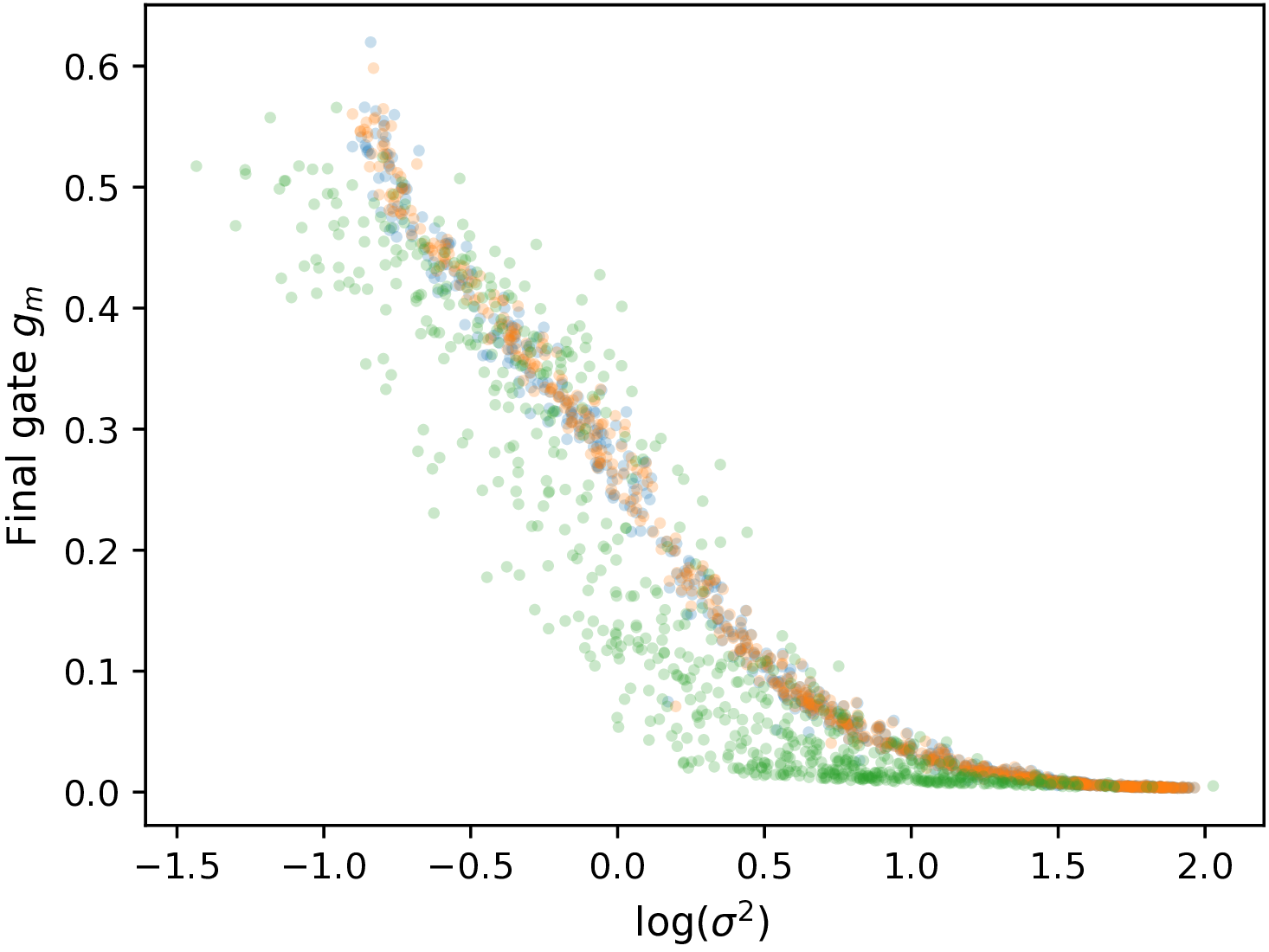}

    (b) Vision
\end{minipage}
\hfill
\begin{minipage}[t]{0.32\textwidth}
    \centering
    \includegraphics[width=0.8\textwidth]{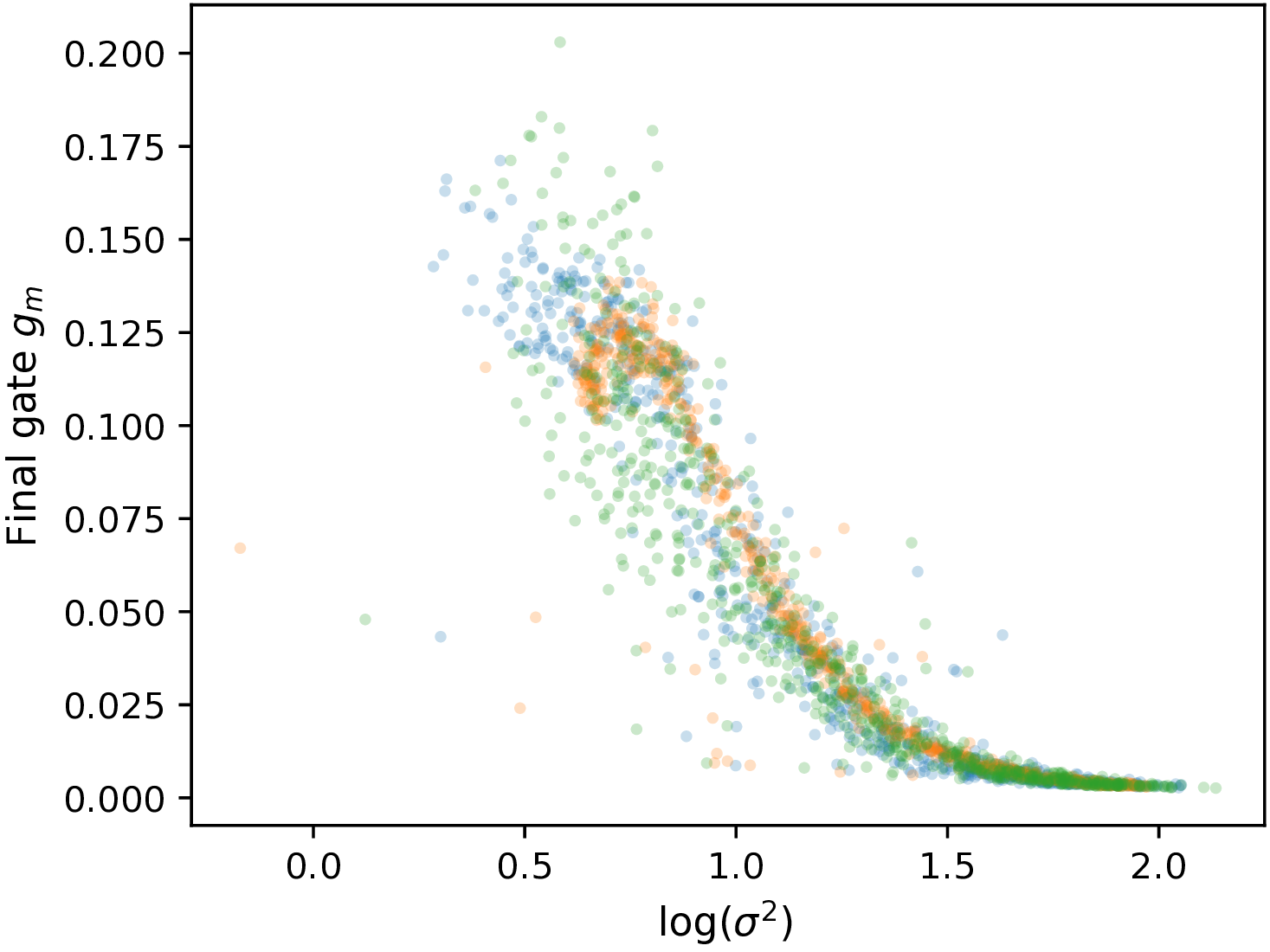}

    (c) Audio
\end{minipage}
\caption{Mechanism analysis of MRUF. Scatter plots of predicted uncertainty $\log(\sigma_m^2)$ versus the final modality gate $g_m$ for language, vision, and audio under clean, drop, and Gaussian perturbation conditions. A negative relationship indicates that modalities with higher predicted uncertainty tend to receive lower fusion weights.}
\label{fig:uaf_router_abc}
\end{figure*}

\begin{table}[t]
\centering
\caption{Ablation of MRUF and MCA on CMU-MOSI (aligned).}
\label{tab:ablation_router_mca}
\footnotesize
\setlength{\tabcolsep}{4pt}
\renewcommand{\arraystretch}{1.05}
\begin{tabular}{c c | cccc}
\hline
MRUF & MCA & ACC$_7$ $\uparrow$ & ACC$_2$ $\uparrow$ & F1 $\uparrow$ & MAE $\downarrow$ \\
\hline
& & 44.5 & 83.2 & 83.3 & 0.729 \\
& $\checkmark$ & 46.7 & 83.6 & 83.6 & 0.734 \\
$\checkmark$ & & 45.1 & 83.8 & 83.7 & 0.718 \\
$\checkmark$ & $\checkmark$ & \textbf{46.7} & \textbf{84.5} & \textbf{84.4} & \textbf{0.717} \\
\hline
\end{tabular}
\end{table}

\begin{table}[t]
\centering
\caption{Ablation on router supervision on CMU-MOSI (aligned).}
\label{tab:router_ablation}
\footnotesize
\setlength{\tabcolsep}{5pt}
\renewcommand{\arraystretch}{1.05}
\begin{tabular}{l|cccc}
\hline
Teacher \& Loss & ACC$_2$ $\uparrow$ & F1 $\uparrow$ & ACC$_7$ $\uparrow$ & MAE $\downarrow$ \\
\hline
none + MSE & 83.0 & 82.9 & 44.2 & 0.731 \\
LOO-drop + KL & 83.9 & 83.9 & 44.5 & 0.729 \\
uniform + CE & 84.4 & 84.3 & 45.2 & \textbf{0.703} \\
sigma-guided + KL & 83.6 & 83.6 & 46.4 & 0.719 \\
LOO-drop + MSE & \textbf{84.5} & \textbf{84.5} & \textbf{46.7} & 0.717 \\
\hline
\end{tabular}
\end{table}

\subsection{Comparison with Representative Baselines}
\label{sec:sota}

We compare MRUF with representative multimodal sentiment analysis baselines, including TFN~\cite{zadeh2017tensor}, LMF~\cite{liu2018efficient}, MulT~\cite{tsai2019multimodal}, RAVEN~\cite{wang2019words}, MICA~\cite{liang2021attention}, MTAG~\cite{yang2021mtag}, PMR~\cite{lv2021progressive}, Self-MM~\cite{yu2021learning}, EMOE~\cite{fang2025emoe}, and DMD~\cite{li2023decoupled}.

\subsubsection{Results on CMU-MOSI}

As shown in Table~\ref{tab:mosi}, MRUF achieves the best overall performance on CMU-MOSI under both aligned and unaligned settings. Compared with DMD, MRUF improves all metrics, indicating that reliability-aware routing and uncertainty-aware gate calibration further strengthen the decoupling-distillation backbone.

\subsubsection{Results on CMU-MOSEI}

Table~\ref{tab:mosei} shows consistent improvements on CMU-MOSEI. Although the gains are more moderate because recent baselines are already competitive on this larger dataset, MRUF still obtains the best overall results under both temporal settings.

\subsection{Ablation Studies}
\label{sec:ablation}

\subsubsection{Effect of MRUF and Modality-Invariant Contrastive Alignment}

Table~\ref{tab:ablation_router_mca} shows that MRUF improves all four metrics over DMD, while MCA mainly strengthens the invariant representation space and improves ACC$_7$. Their combination achieves the best overall performance, confirming their complementary effects.

\subsubsection{Effect of Routing Supervision}

Table~\ref{tab:router_ablation} compares router-supervision strategies under the same architecture. The proposed LOO-drop + MSE strategy achieves the best ACC$_2$, F1, and ACC$_7$, showing that leave-one-out error increases provide effective task-aware supervision.

\subsection{Robustness under Controlled Perturbations}
\label{sec:robustness}

\begin{table}[t]
\centering
\caption{Robustness comparison on CMU-MOSEI (aligned).}
\label{tab:robustness_mosei_aligned}
\footnotesize
\setlength{\tabcolsep}{4pt}
\renewcommand{\arraystretch}{1.05}
\begin{tabular}{l|c|cccc}
\hline
Setting & Model & ACC$_2$ & F1 & ACC$_7$ & MAE \\
\hline
\multirow{2}{*}{Clean} & MRUF & \textbf{84.5} & \textbf{84.4} & \textbf{52.3} & \textbf{0.544} \\
                       & DMD  & 83.5 & 83.5 & 51.6 & 0.553 \\
\hline
\multirow{2}{*}{Text mask 30\%} & MRUF & \textbf{82.3} & \textbf{82.2} & \textbf{51.5} & \textbf{0.586} \\
                                & DMD  & 81.3 & 81.3 & 50.7 & 0.595 \\
\hline
\multirow{2}{*}{Text mask 50\%} & MRUF & \textbf{78.8} & \textbf{78.6} & \textbf{49.9} & 0.677 \\
                                & DMD  & 77.7 & 77.7 & 48.2 & \textbf{0.655} \\
\hline
\multirow{2}{*}{w/o Vision} & MRUF & \textbf{84.1} & \textbf{84.1} & \textbf{51.1} & 0.567 \\
                            & DMD  & 83.2 & 83.1 & 50.2 & \textbf{0.543} \\
\hline
\multirow{2}{*}{w/o Audio} & MRUF & \textbf{84.2} & \textbf{84.3} & \textbf{52.1} & 0.554 \\
                           & DMD  & 83.3 & 83.2 & 51.5 & \textbf{0.538} \\
\hline
\end{tabular}
\end{table}

To examine robustness under degraded inputs, we compare MRUF and DMD under text masking and modality removal. As shown in Table~\ref{tab:robustness_mosei_aligned}, MRUF consistently improves ACC$_2$, F1, and ACC$_7$ across all perturbation settings, indicating stronger categorical robustness when one modality becomes weak or unavailable.

MRUF does not always obtain lower MAE under severe perturbations. This is because uncertainty-aware fusion mainly down-weights unreliable modalities to protect discrete decision boundaries, benefiting ACC$_2$, F1, and ACC$_7$. MAE instead requires fine-grained sentiment intensity estimation. When a modality is heavily masked or removed, such intensity information may be insufficient, even if the model preserves the correct polarity or sentiment interval.

\subsection{Mechanism Analysis of MRUF}
\label{sec:analysis}

MRUF defines the final modality gate as $g_m=g_m^{\mathrm{mod}}\omega_m$, where $\omega_m$ is derived from predicted uncertainty. We analyze the relation between predicted uncertainty and final gates under clean, drop, and Gaussian perturbation conditions.

As shown in Fig.~\ref{fig:uaf_router_abc}, the learned gates decrease as predicted uncertainty increases, especially under degraded inputs. This supports the intended reliability-aware behavior of MRUF.

\section{Conclusion}

This paper presented MRUF, a reliability-aware multimodal fusion method built on a decoupling-distillation backbone. MRUF estimates utterance-level modality importance through task-aware multi-granularity routing and calibrates modality gates with predicted uncertainty to reduce the influence of unreliable modalities. Experiments on CMU-MOSI and CMU-MOSEI demonstrate consistent improvements over strong baselines, while ablation, robustness, and mechanism analyses validate the proposed design. MRUF introduces additional training cost due to leave-one-out router supervision, and regression under severe perturbations remains challenging. Future work will explore lighter router supervision and more efficient uncertainty-aware routing.






\bibliographystyle{IEEEtran}
\bibliography{references}

\end{document}